\documentclass[runningheads]{llncs}
\usepackage[T1]{fontenc}
\usepackage{xcolor} 
\usepackage{graphicx}
\usepackage{booktabs}
\usepackage{listings}
\usepackage{xcolor}
\usepackage{verbatim}
\usepackage{siunitx}
\usepackage{subcaption}
\usepackage{amsmath}
\usepackage{threeparttable}
\usepackage{tikz}
\usetikzlibrary{arrows.meta, positioning, shapes.multipart}
\usepackage[table,xcdraw]{xcolor}

\usepackage{caption}
\usepackage{multirow}
\usepackage{graphicx}

\lstdefinelanguage{json}{
  basicstyle=\ttfamily\small,
  numbers=none,
  showstringspaces=false,
  breaklines=true,
  literate=
   *{0}{{{\color{numb}0}}}{1}
    {1}{{{\color{numb}1}}}{1}
    {2}{{{\color{numb}2}}}{1}
    {3}{{{\color{numb}3}}}{1}
    {4}{{{\color{numb}4}}}{1}
    {5}{{{\color{numb}5}}}{1}
    {6}{{{\color{numb}6}}}{1}
    {7}{{{\color{numb}7}}}{1}
    {8}{{{\color{numb}8}}}{1}
    {9}{{{\color{numb}9}}}{1}
    {:}{{{\color{punct}{:}}}}{1}
    {,}{{{\color{punct}{,}}}}{1},
}

\begin{document}
\title{CitiLink-Minutes: A Multilayer Annotated Dataset of Municipal Meeting Minutes}
\titlerunning{CitiLink-Minutes: Municipal Meeting Minutes Dataset}
%
\author{
Ricardo~Campos\inst{1,3}\orcidID{0000-0002-8767-8126} \and
Ana Filipa Pacheco\inst{2,3}\orcidID{0009-0003-6135-8479} \and
Ana Luísa Fernandes\inst{2,3}\orcidID{0009-0009-0552-3904} \and
Inês Cantante\inst{2,3}\orcidID{0009-0002-3866-4550} \and
Rute Rebouças\inst{2,3}\orcidID{0000-0002-8213-1068} \and
Luís~Filipe~Cunha\inst{3}\orcidID{0000-0003-1365-0080} \and
José~Isidro\inst{2,3}\orcidID{0009-0000-6071-9138} \and
José~Evans\inst{2,3}\orcidID{0009-0003-6408-5103} \and
Miguel~Marques\inst{1,3}\orcidID{0009-0002-7934-0173} \and
Rodrigo~Batista\inst{2,3}\orcidID{0009-0004-0431-6042} \and
Evelin~Amorim\inst{2,3}\orcidID{0000-0003-1343-939X} \and
Alípio~Jorge\inst{2,3}\orcidID{0000-0002-5475-1382} \and
Nuno~Guimarães\inst{2,3}\orcidID{0000-0003-2854-2891} \and
Sérgio~Nunes\inst{2,3}\orcidID{0000-0002-2693-988X} \and
António Leal\inst{2,4}\orcidID{0000-0002-6198-2496} \and
Purificação~Silvano\inst{2,3}\orcidID{0000-0001-8057-5338}
}
\authorrunning{R. Campos et al.}
%
\institute{
University of Beira Interior, Covilhã, Portugal \\
\email{\{ricardo.campos, miguel.alexandre.marques\}@ubi.pt}
\and
University of Porto, Porto, Portugal
\and
INESC TEC, Porto, Portugal 
\email{\{ana.f.pacheco, ana.l.fernandes, ines.cantante, rute.reboucas, luis.f.cunha, jose.m.isidro, jose.joao, rodrigo.f.batista, evelin.f.amorim, alipio.jorge, nuno.r.guimaraes, sergio.nunes, purificacao.silvano\}@inesctec.pt}
\and
University of Macau, Macau 
\email{antonioleal@um.edu.mo}\\
}
\maketitle              
\begin{abstract}

City councils play a crucial role in local governance, directly influencing citizens’ daily lives through decisions made during municipal meetings. These deliberations are formally documented in meeting minutes, which serve as official records of discussions, decisions, and voting outcomes. Despite their importance, municipal meeting records have received little attention in Information Retrieval (IR) and Natural Language Processing (NLP), largely due to the lack of annotated datasets, which ultimately limit the development of computational models. To address this gap, we introduce CitiLink-Minutes, a multilayer dataset of 120 European Portuguese municipal meeting minutes from six municipalities. Unlike prior annotated datasets of parliamentary or video records, CitiLink-Minutes provides multilayer annotations and structured linkage of official written minutes. The dataset contains over one million tokens, with all personal identifiers de-identified. Each minute was manually annotated by two trained annotators and curated by an experienced linguist across three complementary dimensions: (1) metadata, (2) subjects of discussion, and (3) voting outcomes, totaling over 38,000 individual annotations. Released under FAIR principles and accompanied by baseline results on metadata extraction, topic classification, and vote labeling, CitiLink-Minutes demonstrates its potential for downstream NLP and IR tasks, while promoting transparent access to municipal decisions.

\end{abstract}
\setcounter{footnote}{0}
\section{Introduction}
Official city council meeting minutes represent a rich yet underexplored source of civic information. They document local decision-making processes, including policy discussions, proposals, and voting outcomes, offering structured evidence of how local governments operate. Recognizing their importance, several civic data initiatives have emerged to make government proceedings more accessible. Projects such as the Council Data Project~\cite{maxfield2022councils} and citymeetings.nyc\footnote{\url{https://citymeetings.nyc}} provide searchable interfaces and automatic summaries of meeting transcripts from U.S. municipalities. While these resources demonstrate growing interest in the computational analysis of municipal data, they mostly focus on video transcripts~\cite{van2025spoken}, where speech-to-text systems produce transcribed versions of oral deliberation. In contrast, official written meeting minutes are more difficult to analyze, as they are often lengthy and heterogeneous, with structures that vary across municipalities. The fact that relevant information, such as policy topics, proposals, and voting outcomes, is often embedded within long narrative passages makes it challenging for both humans and computational systems to locate, extract, and compare information across meetings and institutions. Consequently, progress in this direction has been slow, as municipal-level data are rarely curated, standardized, or made openly available for computational research. From an IR and NLP perspective, this scarcity limits both model training and evaluation, hindering progress on domain-specific tasks that require resources adapted to the linguistic and structural characteristics of municipal discourse.

To circumvent this limitation, many researchers rely on web-scraped data, which are often noisy and can hamper model training~\cite{penedo2024fineweb}. Such data may also contain sensitive information or raise copyright concerns~\cite{Zeldes2017}, further limiting dataset publication and the advancement of research~\cite{barati2023open}. Existing initiatives also remain largely limited to English and lack the high-quality, multilayer annotations that are essential for deeper linguistic analysis and effective IR research.

To address this gap, we introduce in this paper the CitiLink-Minutes dataset\footnote{\url{https://github.com/INESCTEC/citilink-dataset}} \cite{citilink2025}, developed as part of the CitiLink project\footnote{\url{https://citilink.inesctec.pt/}}. Unlike existing initiatives that primarily focus on video transcripts, CitiLink-Minutes provides official written minutes. The dataset comprises 120 municipal meeting minutes (2021–2024) in European Portuguese, collected in collaboration with six municipalities, ensuring legally compliant and copyright-safe access to official records with consistent, standardized annotations across municipalities. Each meeting minute was manually annotated by a team of linguists following a four-layer schema encompassing personal information (PI), metadata, subjects of discussion, and voting outcomes. The \textbf{PI} layer captures any mention that can identify a person; \textbf{metadata} records details such as date, location, and participants; \textbf{subjects of discussion} encodes topics and themes; and the \textbf{voting} tracks voters and results. A double-annotation process was pursued, followed by curator validation to resolve discrepancies and ensure consistency. To release the data, all PI was manually de-identified to protect privacy and comply with data sensitivity requirements. This effort resulted in a high-quality annotated dataset, richly structured and multilayer. It addresses privacy and copyright concerns and offers linguistic and structural depth unmatched by existing municipal datasets. We also provide an interactive dashboard\footnote{\url{https://dataset.citilink.inesctec.pt} To access the platform, visit INESC TEC repository (https://doi.org/10.25747/7KG6-1K22) and request access to the dataset} that allows users to explore the dataset in detail.

The rich annotations enable a variety of tasks, including metadata extraction, topic classification, subject detection, and vote labeling. To support reproducible research and provide reference points, we offer baseline results, highlighting both the opportunities and challenges of working with municipal meeting minutes in downstream tasks. Our contributions are summarized as follows.

\begin{enumerate}
\item A novel, human-annotated dataset of 120 municipal minutes from six Portuguese municipalities, with dense annotations across four layers: PI, meeting metadata, subjects of discussion, and voting outcomes. The dataset is released under a permissive license following FAIR principles, with all personal data de-identified to ensure privacy, guaranteeing unrestricted access.

\item An interactive dashboard for detailed exploration of the dataset, allowing users to inspect its structure and annotations.

\item The definition of two redefined tasks (voter identification, topic classification) and one novel task (metadata identification) for municipal meeting minutes, along with evaluation metrics and baselines (made available on Hugging Face) that establish reference points and ilustrate the complexity of the dataset, guiding future research.

\end{enumerate}

The remainder of this work is organized as follows. Section \ref{sec:relatedwork} reviews related work. Section \ref{sec:dataset} describes the dataset, its statistics, and the annotation scheme. Section \ref{sec:experiments} presents the baseline results for a selection of tasks supported by the dataset. Finally, Section \ref{sec:conclusion} summarizes the main findings, followed by the discussion of limitations.

\section{Related Work}\label{sec:relatedwork}

Research on institutional and governmental discourse has largely relied on parliamentary proceedings, which primarily capture spoken interaction. Well-known examples include EuroParl~\cite{koehn2005europarl}, a multilingual parallel corpus of the official proceedings of the European Parliament, widely used in machine translation, the European Parliament debates~\cite{vanAggelen2016}, and the ParlaMint II project~\cite{ParlaMintII}, which provides richly annotated, multilingual parliamentary corpora from 29 European countries. Other datasets have introduced finer-grained annotations. For instance, Mor-Lan et al.~\cite{mor-lan-etal-2024-israparltweet} annotated Israeli parliamentary debates for member identification and session metadata, leveraging the consistent formatting of parliamentary records, a regularity far less common in city council minutes, which makes metadata annotation particularly challenging. Other studies have focused on speech acts~\cite{reinig-etal-2024-politics}, speaker attribution~\cite{rehbein-etal-2024-mouths}, and causality~\cite{garcia-corral-etal-2024-politicause}, covering multiple languages ~\cite{frasnelli-palmero-aprosio-2024-theres,mor-lan-etal-2024-israparltweet,yrjanainen-etal-2024-swedish}, including Brazilian Portuguese~\cite{fernandes2024publichearingbr}, and European Portuguese~\cite{albertina}. These resources demonstrate strong focus on parliamentary debate, with high-quality annotations, metadata, and structured formats.

In contrast, municipal governance corpora remain scarce, offering only limited opportunities to advance narrative understanding~\cite{santana2023}. Existing resources mostly rely on audio or video transcriptions with limited annotations. For example, LocalView~\cite{barari2023localview} provides video transcripts of U.S. local meetings, yet lacks linguistic annotations. Similarly, van Wijk and Marx~\cite{van2025spoken} offer a small dataset of transcribed Dutch municipal meetings, also without annotations, while Spangher et al.~\cite{spangher2023tracking} annotated San Francisco city council meeting transcriptions with media coverage labels. Building on similar objectives, MeetingBank~\cite{Hu2023MeetingBankAB} provides segmented city council video transcripts for summarization, while Ulysses~\cite{albuquerque2022ulyssesner,siqueira2024segmenting}, a dataset of Brazilian legislative bills, offers segmentation and NER annotations in Brazilian legislative and municipal contexts, but focuses on general entities, missing the full richness of municipal meeting content. Among these efforts, only Vlantis et al.~\cite{vlantis-etal-2024-benchmarking} use municipal textual documents, but with limited scope and annotation. To address these limitations, we propose the CitiLink-Minutes dataset, a comprehensive, multilayer-annotated corpus of 120 municipal minutes in European Portuguese, from six Portuguese municipalities.

\section{The CitiLink-Minutes Dataset}\label{sec:dataset}
Here we present the CitiLink-Minutes dataset, providing a comprehensive account of its construction process, the underlying annotation scheme, and the principal characteristics of the resulting annotated corpus.

\subsection{Dataset Creation}\label{datasetcreation}

The meeting minutes were collected under partnership agreements with six municipalities - Alandroal, Campo Maior, Covilhã, Fundão, Guimarães and Porto - granting access to all official records produced during the 2021–2024 administrative term. In total, 479 minutes were obtained. From this collection, the linguistics team selected a representative subset for annotation based on predefined selection and processing criteria:
(i) only minutes dated from October 2021 onwards were included, corresponding to the start of the term of office of the mayor;
(ii) a diversity of session types, such as ordinary, extraordinary, public, and private, was ensured;
(iii) meetings containing between 15 and 40 agenda items were selected to balance topic representativeness with annotation feasibility; and
(iv) public meetings were preferred, since they are open to citizens and the media and typically address matters of broader civic relevance.
Applying these criteria yielded a final selection of 120 minutes, 20 per municipality. Table~\ref{tab:minutes-annual} presents the annual distribution of the collected minutes across the six municipalities, with the selected subset indicated in parentheses.

\begin{table}[t]
\centering
\caption{Number of meeting minutes collected and (selected) per year and municipality (2021–2024).}
\label{tab:minutes-annual}
\small
\setlength{\tabcolsep}{3.5pt}
\renewcommand{\arraystretch}{1.05}
\resizebox{\columnwidth}{!}{%
\begin{tabular}{lrrrrrrr}
\toprule
\textbf{Year} & \textbf{Alandroal} & \textbf{Campo Maior} & \textbf{Covilhã} & 
\textbf{Fundão} & \textbf{Guimarães} & \textbf{Porto} & \textbf{Total} \\
\midrule
2021 & 7 (2) & 25 (2) & 21 (2) & 4 (2) & 5 (2) & 6 (2) & 68 (12) \\
2022 & 27 (6) & 25 (6) & 20 (6) & 16 (6) & 22 (6) & 24 (6) & 134 (36) \\
2023 & 28 (6) & 25 (6) & 21 (6) & 17 (6) & 22 (6) & 24 (6) & 137 (36) \\
2024 & 30 (6) & 25 (6) & 21 (6) & 18 (6) & 23 (6) & 23 (6) & 140 (36) \\
\midrule
\textbf{Total} & \textbf{92 (20)} & \textbf{100 (20)} & \textbf{83 (20)} &
\textbf{55 (20)} & \textbf{72 (20)} & \textbf{77 (20)} & \textbf{479 (120)} \\
\bottomrule
\end{tabular}%
}
\end{table}

\subsection{Annotation Scheme and Methodology}\label{annotationscheme}

The annotation framework developed for this dataset follows the modular architecture proposed by ISO 24617: Language Resource Management - Semantic Annotation Framework (SemAF, e.g.~\cite{ISO24617-6}), and comprises two main types of structures, \textbf{Entities} (\(E\)) and \textbf{Links} (\(\mathcal{L}\)). While grounded in the ISO 24617 specification, our framework adopts a broader interpretation of semantic information, allowing for richer, context-sensitive, and task-specific annotations.

Formally, the set of \textbf{Entities}, is defined as \(E = \{ e_1, \ldots, e_{n_E} \}\), where each entity \( e_i \) is modeled as \(e_i = (m, s, a_e, v_e)\). Here \(m\) represents a \textit{markable} (i.e., a text span identified in the document), \(s\) is the semantic label assigned to that span, \(a_e\) is an optional attribute associated with \(s\) drawn from the set \(A_e(s)\), and \(v_e\) is an optional value associated with \(a_e\) drawn from the set \(V_e(a_e)\). For example, an entity may correspond to the text span "Rui Moreira", annotated with the semantic label \(s = \textit{Participant}\), the attribute \(a_e = \textit{Mayor}\), and the value regarding political party \(v_e = \textit{RM}\). In this case, the entity is represented as \(e = (\text{"Rui Moreira"}, \textit{Participant}, \textit{Mayor}, \textit{RM})\).

Formally, the set of \textbf{Links} is defined as  \(\mathcal{L} = \{ l_1, \ldots, l_{n_L} \}\), where each link \( l_i \) is modeled as \(l_i = (e_p, e_q, r, a_l)\). Here, \(e_p, e_q \in E\) are the two entities connected by the link, \(r\) represents the \textit{relation type} between them, and \(a_l\) is an optional attribute associated with the relation \(r\) drawn from the set \(A_l(r)\). For example, a link of type \(\textit{Voting}\) may connect the entities $e_p = $ \text{("Rui Moreira"}, \textit{Participant}, \textit{Mayor}, \textit{RM}) and \(e_q = (\text{"deliberated"}, \textit{voting\_evidence})\), with the attribute \(a_l = \textit{in favour}\). In this case, the link is represented as \(l = (e_p, e_q, \textit{Voting}, \textit{in favour})\).

\begin{figure}[htbp]
    \centering
    \begin{subfigure}[b]{0.29\textwidth}
        \centering
        \includegraphics[width=\textwidth]{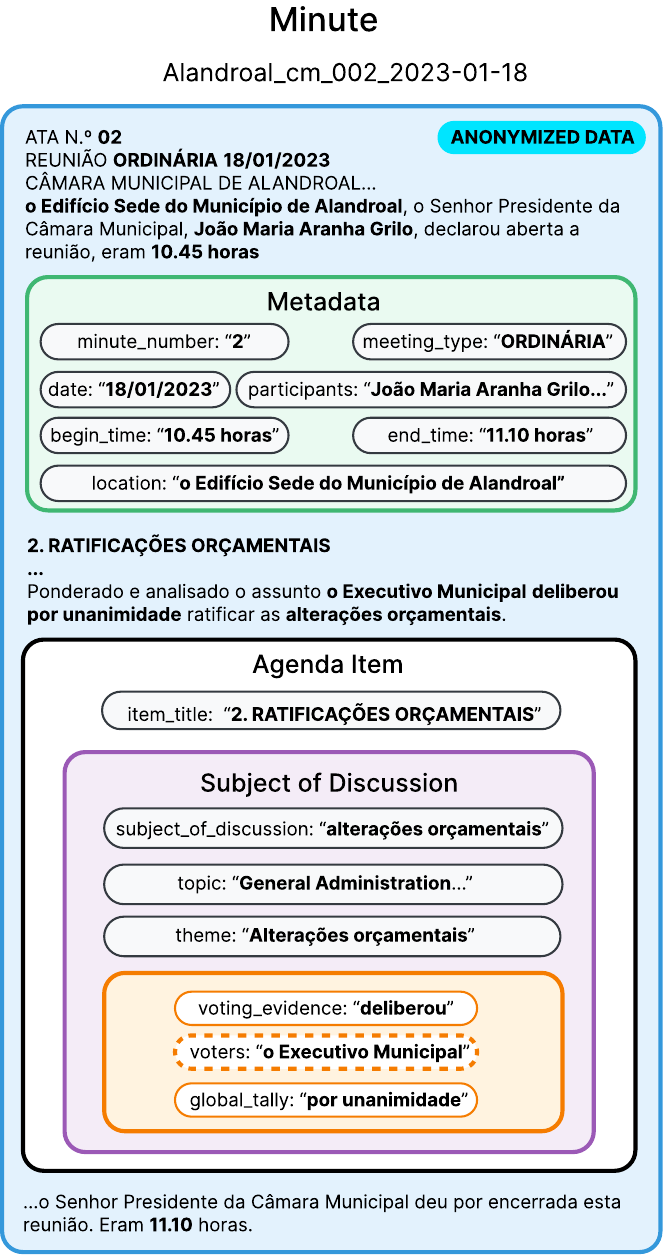}
        \caption{Representative instance of an annotated minute.}
        \label{fig:minute-diagram}
    \end{subfigure}
    \hspace{0.02\textwidth}
    \begin{subfigure}[b]{0.60\textwidth}
        \centering
        \includegraphics[width=\textwidth]{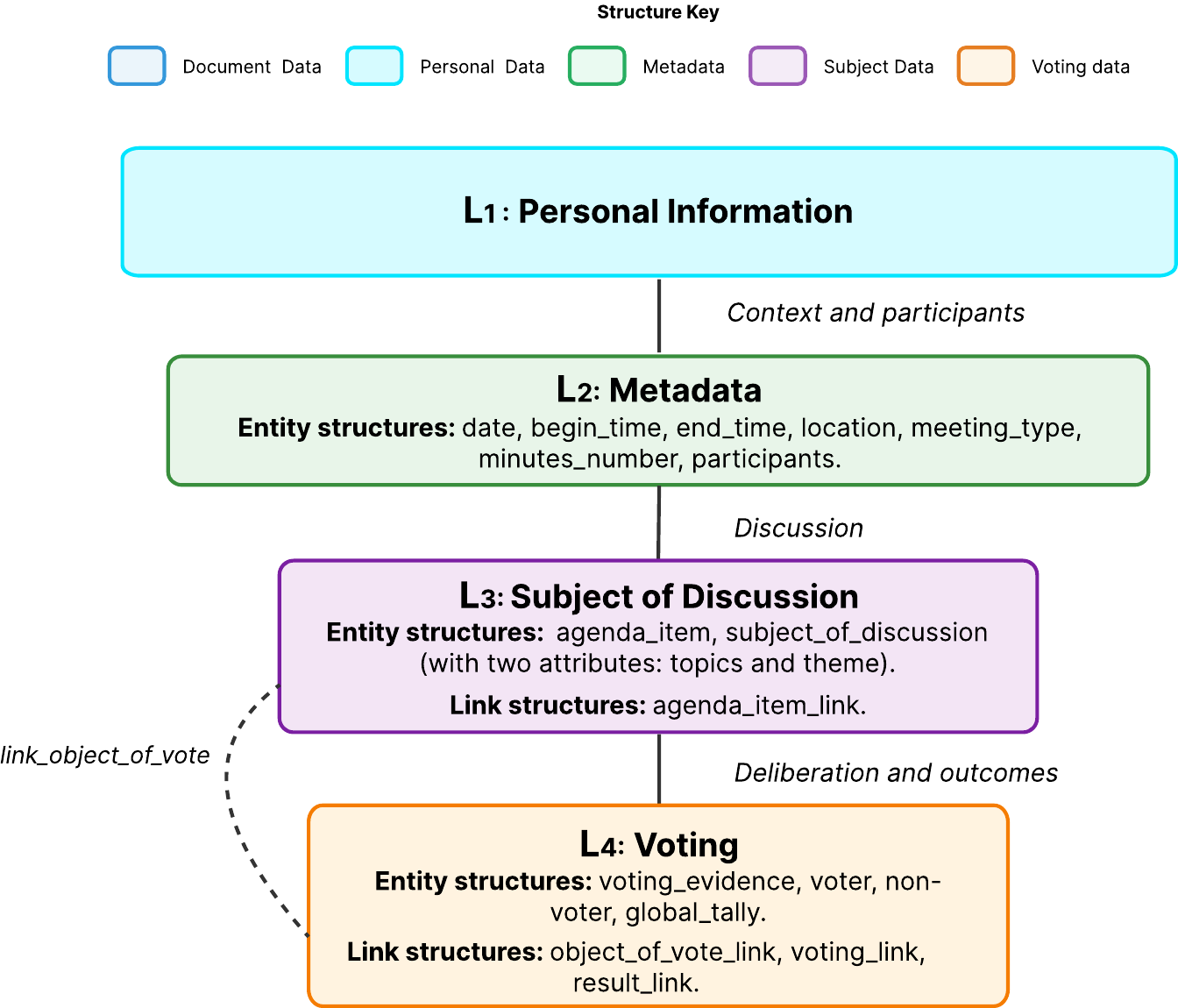}
        \caption{Schematic representation of the four-layer annotation framework.}
        \label{fig:annotation-framework-sub}
    \end{subfigure}
    \caption{Overview of the proposed framework and its implementation.}
    \label{fig:annotation-framework}
\end{figure}

Together, these two types of structures are applied across four conceptual annotation layers (\(L_1, L_2, L_3, L_4\)). The first layer, \(L_1\): \textbf{Personal Information}, captures information that can directly or indirectly identify a person, such as proper names, addresses, etc. The second layer, \(L_2\): \textbf{Metadata}, includes entities representing general information about the meeting, for instance participants, dates, locations, and times. The third layer, \(L_3\): \textbf{Subject of discussion}, represents the main topics and themes of the deliberations. Finally, the fourth layer, \(L_4\): \textbf{Voting}, captures the outcomes of votes on each subject, including the voters, non-voters, voting evidence, and global tally. Figure \ref{fig:annotation-framework}a presents a representative annotated minute, illustrating the information captured by the annotation framework. Figure \ref{fig:annotation-framework}b schematically summarizes the four annotation layers.

Four linguistics students (ANN1–ANN4) were responsible for the annotation and curation of the meeting minutes that constitute the corpus. To ensure consistency and resolve ambiguities, weekly meetings were held with two senior researchers experienced in semantic annotation and large annotation projects such as SemEval and CLEF \cite{piskorski-etal-2025-semeval,JRC138585}. Before the large-scale annotation, conducted on the INCEpTION platform~\cite{klie2018inception}, a pilot phase was carried out involving ten minutes per municipality. This phase aimed at training the annotators and also served to assess whether LLM-based metadata pre-annotation could enhance the speed and efficiency of the process. The results indicated that this approach reduced the time required for metadata annotation by roughly two hours per document, which is in line with other works \cite{cunha_2025,Li_2023} that have shown the utility of LLMs in improving annotation workflows through human-AI collaboration. A detailed analysis of this experiment lies beyond the scope of this paper. All four annotators jointly annotated one minute from each municipality to resolve ambiguities, align interpretations of the annotation guidelines, and establish a common understanding for subsequent individual work. 
To ensure annotation quality~\cite{Biemann2017,snow-etal-2008-cheap}, each minute was annotated across all layers by two annotators (\textit{ANN1}/\textit{ANN2} or \textit{ANN3}/\textit{ANN4}) and subsequently validated by a curator. The only exception was the \(L_2\): \textit{Metadata} layer, which was annotated by a single annotator due to the straightforward nature of the task, though still subject to curator validation. In cases of ambiguity or uncertainty, the curator consulted the responsible annotators and domain experts, and, when appropriate, initiated a collective deliberation to reach a consensus~\cite{Oortwijn2021}.

\subsection{Inter-Annotator Agreement (IAA)}\label{IAA}
\begin{figure}[htbp]
\includegraphics[width=\textwidth]{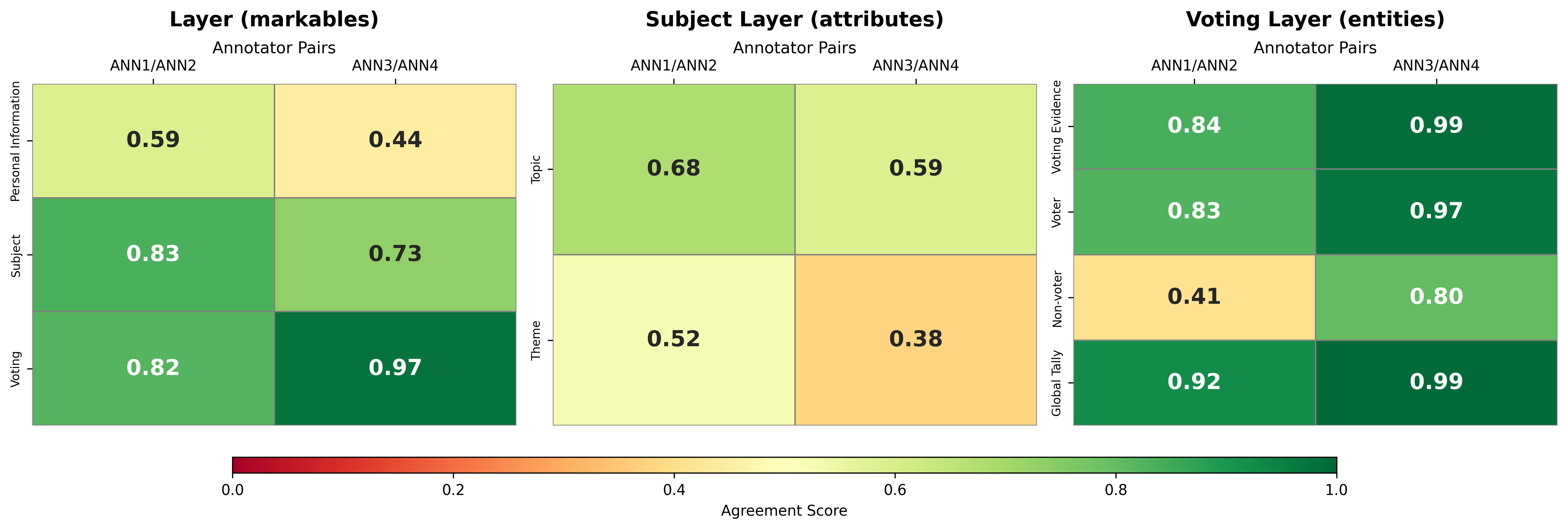}

    \caption{Inter-annotator Agreement scores across different annotation layers.}
    \label{fig:agreement-comparison}
\end{figure}

To assess the reliability of our annotations across different layers and categories, we measured Inter-Annotator Agreement (IAA) using Krippendorff’s alpha~\cite{artstein2008inter}. Figure \ref{fig:agreement-comparison} summarizes the agreement values between annotator pairs across all layers and annotation types. 

Overall, the IAA values indicate consistent agreement across most annotation layers and categories, with some exceptions (Figure \ref{fig:agreement-comparison}). The leftmost chart presents overall agreement on markables, by layer, independent of specific entity or attribute assignments. In the \textit{personal information} layer, spans were considered matching if annotators differed by at most one word. Nonetheless, agreement remains moderate, likely due to the lack of clearly defined entity types (e.g., occupations, administrative information) in this layer, which are still under development and expected to improve annotation precision in the future. The \textit{subject} and \textit{voting} layers show generally strong agreement between annotator pairs. However, agreement for the subject's attributes, \textit{theme} and \textit{topic}, is lower, as illustrated in the middle heatmap. The reduced agreement for \textit{topic} reflects the large number of possible values (n = 22) and the fact that a subject may legitimately correspond to multiple topics. The \textit{theme} attribute shows the lowest agreement, likely due to its open-text format, where annotators independently described the thematic content of each discussion subject. To assess agreement fairly, annotations were considered matching if they achieved at least 70\% similarity according to the Levenshtein distance (i.e., the number of character-level edits required to transform one string into another), and Krippendorff's alpha was then computed based on these matches. In the \textit{voting} layer, the only exception occurred for the \textit{non-voter} entity, where agreement between ANN1 and ANN2 was 0.41. This discrepancy can be attributed to a misinterpretation by ANN2 regarding that specific entity. However, this was fixed during the curation phase.

\subsection{Structure and Dataset Characterization}\label{datasetchracterization}

\begin{figure}[t]
    \centering
    \includegraphics[width=0.9\linewidth]{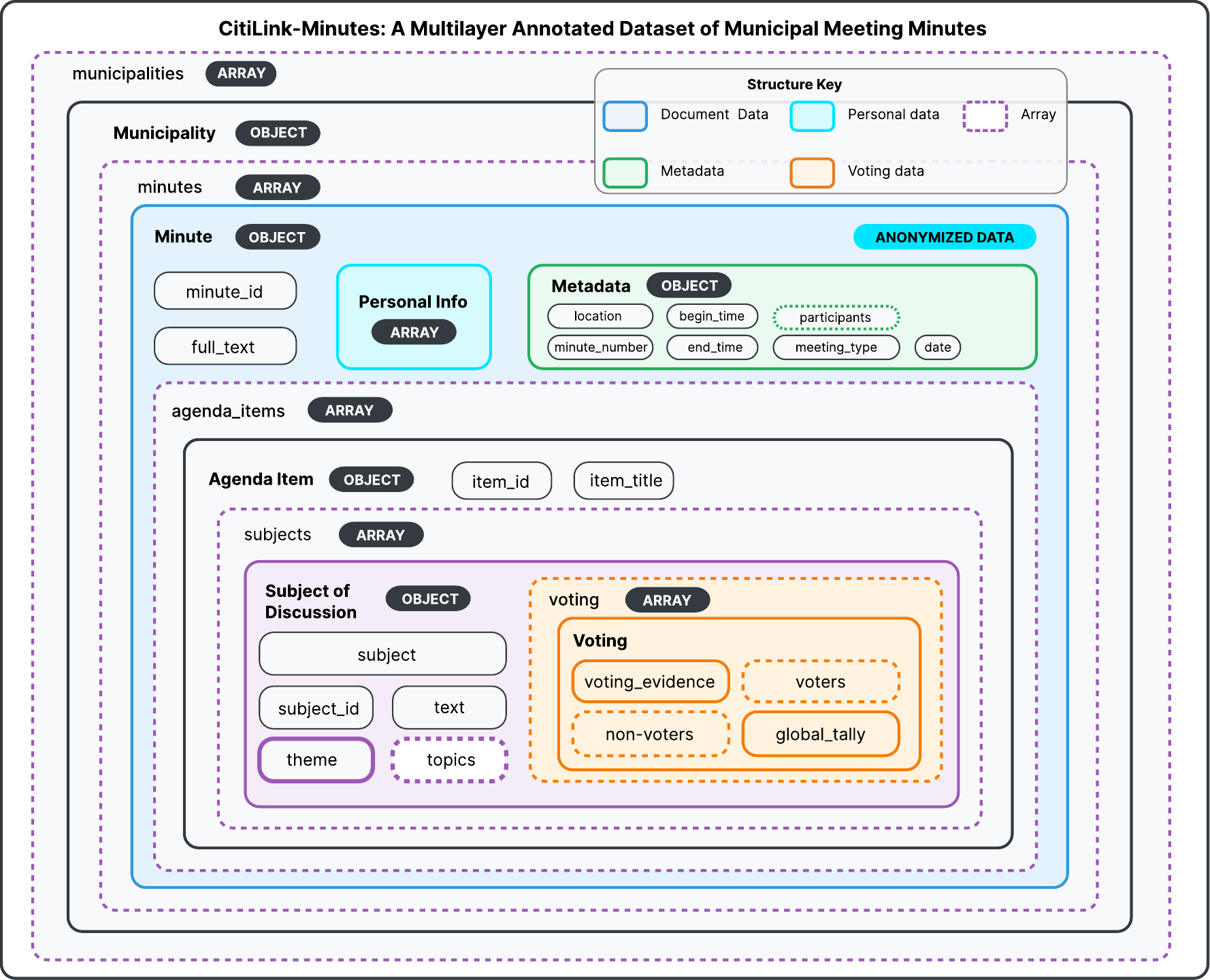}
    \caption{Summary of the JSON structure of CitiLink-Minutes dataset.}
    \label{fig:dataset_stucture}
\end{figure}

The CitiLink-Minutes dataset comprises 120 meeting minutes, providing a robust resource for research on IR and NLP in municipal governance. While the number of minutes is moderate, each document is richly annotated across multiple layers, making it a valuable pioneering resource in European Portuguese and establishing a benchmark for future research. The dataset is organized into six JSON files, one per municipality, each containing 20 annotated documents. Figure~\ref{fig:dataset_stucture} shows the structure of the JSON files. Each document contains the four annotation layers: Personal Information, Metadata, Subject of Discussion, and Voting. The dataset is publicly available on GitHub (see footnote 1) under the Creative Commons Attribution–NonCommercial–NoDerivatives 4.0 International License (CC-BY-NC-ND), and on the INESC TEC repository with a persistent DOI~\cite{citilink2025}, ensuring broad access and reuse within the research community. In addition to the full dataset, we define a data split comprising 72 meeting minutes for training, 24 for validation, and 24 for testing. The documents were ordered chronologically, with the most recent minutes reserved for the test set. This temporal split ensures that evaluation reflects a realistic deployment scenario, where models are tested on newer, unseen data that simulates future meetings.

Table \ref{tab:generalstats} summarizes the statistics of the CitiLink-Minutes dataset across six municipalities, comprising over 1 million tokens, with a total of 20,375 entities and 11,162 relations. Covilhã shows the highest number of relations (2,585) and entities (4,518), suggesting detailed meeting descriptions and richer inter-entity connections. In contrast, Fundão has a smaller set of entities (1,915) and relations (983) despite a relatively high token count ($\approx$189 K). Campo Maior and Porto display moderate document sizes ($\approx$150–160K tokens) but maintain a relatively high entity density, while Alandroal ($\approx$52 K tokens) still contains nearly 3K entities, showing that even shorter minutes can be semantically dense. Overall, the corpus exhibits substantial content and structural diversity across municipalities, which is valuable for evaluating model robustness.

\begin{table}[t]
\centering
\caption{Statistics for the CitiLink\mbox{-}Minutes dataset.}
\label{tab:generalstats}
\small
\setlength{\tabcolsep}{3.8pt}
\renewcommand{\arraystretch}{1.08}
\resizebox{0.5\columnwidth}{!}{%
\begin{tabular}{lrrr}
\toprule
\textbf{Municipality} &
\textbf{Tokens} &
\textbf{Entities} &
\textbf{Relations} \\
\midrule
Alandroal   & 51,987  & 2,902 & 1,796 \\
Campo Maior & 161,889 & 4,187 & 1,474 \\
Covilhã     & 235,381 & 4,518 & 2,585 \\
Fundão      & 189,128 & 1,915 &   983 \\
Guimarães   & 206,361 & 3,547 & 2,154 \\
Porto       & 151,766 & 3,306 & 2,170 \\
\midrule
\textbf{Total} & \textbf{1,016,825} & \textbf{20,375} & \textbf{11,162} \\
\bottomrule
\end{tabular}%
}
\end{table}

Following, Table \ref{tab:layerstats} presents the statistics across the four annotation layers: personal information, metadata, subjects of discussion, and voting. In the \textbf{personal info} layer, we report all the instances that directly or indirectly identify an individual. For further details on the annotation process and specific labeling criteria, we refer the reader to the annotation guidelines manual. For the \textbf{metadata} layer, we focus on the entity \textit{participants}, which aggregates mentions of  \textit{councilors}, and \textit{staff} across the 120 meetings. The entities \textit{president} and \textit{vice-president} are excluded, as there is only one of each per municipality. The remaining metadata entities (\textit{municipality}, \textit{start\_time}, \textit{end\_time}, \textit{minute\_number}, \textit{location}, \textit{meeting\_type}, and \textit{date}) are directly tied to the number of documents, \emph{i.e.}, 20 per municipality. In the \textbf{subjects of discussion} layer, we identify a total of 2,848 subjects across all municipalities, averaging 23.73 per minute. Each subject is associated with one \textit{theme} (thus matching the number of subjects) and one or more \textit{topics}, with an average of 1.7 topics per subject. Covilhã shows the highest number of discussed subjects ($sub_{total}$). Of all identified subjects, 2,489 involve at least one voting procedure ($sub_{vot}$). Finally, the \textbf{voting} layer includes the entities \textit{voting evidence}, \textit{voters}, \textit{non-voters}, and \textit{global tally}. A total of 2,716 voting records were identified (some $sub_{vot}$ include multiple votes), corresponding to an average of 22.63 votes per minute. The entities \textit{voters} and \textit{non-voters} represent aggregated counts of councilor activity per subject rather than individual records. As expected, municipalities with a higher number of discussed subjects exhibit higher numbers of voting entries. For example, Covilhã has the highest number of voting evidence (753) and voters (1,319), reflecting its larger number of subjects. The number of \textit{non-voters} remains relatively low across municipalities, with Guimarães showing the highest count (51), indicating occasional participants who, for different reasons (like incompatibilities), do not participate in a particular voting event.
Building on this aggregated voting information, we also analyzed the distribution of voting positions \textit{in favor}, \textit{against}, or \textit{abstention} for each subject. Overall, the majority of explicit voting positions are \textit{in favor} (67.6\%), while Covilhã shows a higher proportion of \textit{abstentions} (53.7\%), and Porto records the largest share of votes \textit{against} (13.2\%). These patterns highlight both the variation in council decision-making styles and the relationship between discussion volume and voting activity across municipalities.

\begin{table}[t]
    \centering
    \captionsetup{font=footnotesize, skip=2pt}
    \caption{Entities and attributes statistics across layers.}
    \setlength{\tabcolsep}{3.8pt}
    \renewcommand{\arraystretch}{1.08}
    \resizebox{\columnwidth}{!}{%
    \begin{tabular}{l|r|rr|rrr|rrrr}
    \toprule
        \multicolumn{1}{c|}{\textbf{Municipality}} &
        \multicolumn{1}{c|}{\textbf{Personal Info}} &
        \multicolumn{2}{c|}{\textbf{Metadata}} &
        \multicolumn{3}{c|}{\textbf{Subjects}} &
        \multicolumn{4}{c}{\textbf{Voting}} \\ 
        \cmidrule(lr){3-4}\cmidrule(lr){5-7}\cmidrule(lr){8-11}
        & & \textbf{Councilors} & \textbf{Staff} & \textbf{Subj. Total} & \textbf{Subj. Votes} & \textbf{Topics} & \textbf{V. Evidence} & \textbf{Voters} & \textbf{Non-voters} & \textbf{Global tally} \\ 
        \midrule
        Alandroal & 458 & 60 & 20 & 503 & 404 & 736 & 404 & 485 & 10 & 404 \\
        Campo Maior & 2,297 & 60 & 42 & 396 & 328 & 638 & 362 & 393 & 5 & 357 \\
        Covilhã & 1,273 & 117 & 27 & 688 & 576 & 1090 & 753 & 1319 & 23 & 2 \\
        Fundão & 414 & 100 & 20 & 235 & 234 & 459 & 234 & 282 & 3 & 232 \\
        Guimarães & 928 & 188 & 18 & 554 & 486 & 998 & 494 & 628 & 51 & 486 \\
        Porto & 244 & 222 & 22 & 472 & 461 & 891 & 469 & 770 & 8 & 467 \\
        \hline
        \textbf{Total} & \textbf{5,614} & \textbf{747} & \textbf{149} & \textbf{2,848} & \textbf{2,489} & \textbf{4,812} & \textbf{2,716} & \textbf{3,877} & \textbf{100} & \textbf{1,948} \\
    \bottomrule
    \end{tabular}
    }
    \label{tab:layerstats}
\end{table}


\section{Tasks and Baseline Results}\label{sec:experiments}
To demonstrate the potential of the CitiLink-Minutes dataset, we define three representative NLP tasks and establish baseline results for each, covering distinct application scenarios: structured information extraction, entity recognition, and multi-label classification. These experiments are not exhaustive model evaluations but rather reproducible baselines that provide reference points for future research. For each task, we compare an encoder-based and a generative model (few-shot), reporting standard evaluation metrics consistently at the macro (ma) and micro (mi) levels. All experiments followed the predefined data split provided with this resource paper.  The encoder-based baselines (\texttt{BERTimbau Large}~\cite{souza2020bertimbau}) were fine-tuned using \texttt{AdamW} with a learning rate of $5 \times 10^{-5}$, batch size 16, weight decay of 0.01, and a maximum sequence length of 512. Training was run for 10 epochs, with a 10\% warmup phase. For span-based extraction tasks (\textit{Voting Identification} and \textit{Metadata Extraction}), a standard token-classification head was used, while \textit{Topic Classification} utilized a multi-label output layer trained with binary cross-entropy loss (\texttt{BCEWithLogitsLoss}). For the generative baselines, we employed a few-shot prompting strategy with task-specific examples. Each prompt was structured into three components: (i) a task definition including specific constraints, such as the requirement for exact span extraction for \textit{Metadata} and \textit{Voting Identification}, or a predefined label list for \textit{Topic Classification};(ii) the examples illustrating the expected input--output mapping; and (iii) the target inference instance. To ensure consistency, prompts for extraction tasks enforced rules against paraphrasing and the use of exact offsets, while classification prompts required the use of the provided taxonomy. The baseline models are publicly available on HuggingFace\footnote{https://huggingface.co/collections/liaad/citilink-68f7916f31b9588c4fe2f43b} along with the corresponding generative prompts, which are released on the resource paper GitHub repository to support reproducibility and future benchmarking.

\paragraph{\textbf{Metadata Identification}}
The first task focuses on recognizing eight entity types defined in the \(L_2\) \textbf{Metadata} layer within meeting minute segments. We fine-tuned a BERTimbau encoder model as our baseline, while Gemini-2.5-Pro served as the generative counterpart, integrated through the LangExtract framework~\cite{langextract}. Evaluation was conducted using Precision, Recall, and F1-score. As shown in Table~\ref{tab:metadata_vote}, the fine-tuned transformer achieved consistently strong results across metrics, while the generative model exhibited higher precision but substantially lower recall. The narrow gap between macro and micro F1-scores suggests balanced performance across entity types. These findings demonstrate the contrasting behaviors of encoder-based and generative models in structured information extraction, highlighting the strengths of the former in handling domain-specific structured text and establishing them as the primary baseline for future research in this task.

\paragraph{\textbf{Voting Identification}}
The second task focuses on vote identification, aimed at extracting text spans corresponding to voting-related entities: \textit{Voter-Favor}, \textit{Voter-Against}, \textit{Voter-Abstention}, \textit{Voter-Absent}, \textit{Voting}, \textit{Subject}, \textit{Counting-Unanimity}, and \textit{Counting-Majority}, from annotations in the \(L_4\) \textbf{Voting} layer. We fine-tuned a BERTimbau model as the encoder baseline and compared it with Gemini-2.5-Pro integrated into the LangExtract framework~\cite{langextract} for generative span extraction. Performance was measured with Precision, Recall, F1-score (Table~\ref{tab:metadata_vote}). The fine-tuned BERTimbau model achieved robust results across most entities, confirming its effectiveness for structured extraction. The gap between macro and micro F1-scores shows uneven performance across entity types: frequent categories such as \textit{Voting} and \textit{Voter-Favor} are recognized more accurately, while less frequent or lexically variable entities like \textit{Subject} are more challenging. The generative model achieved moderate precision and recall, resulting in lower overall F1-scores compared to BERTimbau. These findings suggest that encoder-based approaches provide more comprehensive coverage for structured vote identification, establishing a stronger baseline than generative models.

\begin{table}[h!]
\centering
\captionsetup{font=small, skip=2pt}
\caption{Evaluation results on Metadata and Vote Identification tasks.}
\label{tab:metadata_vote}
\begin{threeparttable}
\resizebox{0.8\linewidth}{!}{
\renewcommand{\arraystretch}{0.9}
\begin{tabular}{ll@{\hspace{1em}}c@{\hspace{1em}}c@{\hspace{1em}}c@{\hspace{1em}}c@{\hspace{1em}}c@{\hspace{1em}}c}
\toprule
\textbf{Task} & \textbf{Model} &
$\mathbf{F1_{\text{ma}}}$ & $\mathbf{F1_{\text{mi}}}$ &
$\mathbf{P_{\text{ma}}}$ & $\mathbf{P_{\text{mi}}}$ &
$\mathbf{R_{\text{ma}}}$ & $\mathbf{R_{\text{mi}}}$ \\
\midrule
\multirow{2}{*}{\textbf{Metadata}} 
& BERTimbau & 0.752 & 0.959 & 0.743 & 0.958 & 0.771 & 0.960 \\
& Gemini-2.5-Pro & 0.273 & 0.273 & 0.833 & 0.833 & 0.163 & 0.163 \\
\midrule
\multirow{2}{*}{\textbf{Vote}} 
& BERTimbau & 0.814 & 0.705 & 0.750 & 0.577 & 0.927 & 0.904 \\
& Gemini-2.5-Pro & 0.556 & 0.584 & 0.527 & 0.529 & 0.596 & 0.651 \\
\bottomrule
\end{tabular}
}
\end{threeparttable}
\end{table}

\paragraph{\textbf{Topic Classification}}
The third task addresses multi-label topic classification, where each discussion subject (\(L_3\): \textbf{Subject of Discussion}) is assigned one or more labels from 22 predefined topics (1.69 topics per subject on average). This task is inherently imbalanced and semantically diverse. Both an encoder-based BERTimbau model and Gemini-2.5-Pro were evaluated. Performance was assessed using standard multi-label metrics: macro- and micro-F1 (balanced and overall performance), Hamming Loss (HL; lower is better), and Average Precision (AP). As shown in Table~\ref{tab:topics}, the fine-tuned encoder achieved substantially stronger results across all metrics. Overall, the fine-tuned encoder demonstrated its capacity to handle multi-label settings and class imbalance more effectively than the generative baseline. These findings establish a solid foundation for future experimentation on topic modeling and classification over municipal meeting records.

\begin{table}[h!]
\centering
\caption{Evaluation results on the Multi-label Topic Classification task.}
\label{tab:topics}
\begin{threeparttable}
\resizebox{0.8\linewidth}{!}{ 
\renewcommand{\arraystretch}{0.9} 
\begin{tabular}{ll@{\hspace{1em}}c@{\hspace{1em}}c@{\hspace{1em}}c@{\hspace{1em}}c@{\hspace{1em}}c}
\toprule
\textbf{Task} & \textbf{Model} &
$\mathbf{F1_{\text{ma}}}$ &
$\mathbf{F1_{\text{mi}}}$ &
$\mathbf{HL}$ &
$\mathbf{AP_{\text{mi}}}$ &
$\mathbf{AP_{\text{ma}}}$ \\
\midrule
\multirow{2}{*}{\textbf{Topic~Class.}} 
& BERTimbau & 0.642 & 0.822 & 0.029 & 0.870 & 0.681 \\
& Gemini-2.5-Pro & 0.496 & 0.525 & 0.070 & 0.317 & 0.396 \\
\bottomrule
\end{tabular}
}
\end{threeparttable}
\end{table}

\section{Conclusion}\label{sec:conclusion}
In this paper, we introduced CitiLink-Minutes, a novel corpus of 120 densely human-annotated municipal meeting minutes in European Portuguese, created through a partnership with six Portuguese municipalities. The dataset contains high-quality annotations across four layers: personal information, metadata, subjects of discussion, and voting, making it the first resource of its kind to cover all these aspects in municipal records. Its richness and scope support a wide range of downstream NLP and IR tasks, including structured information extraction, vote identification, and multi-label topic classification, and provide reliable baselines for future research. Overall, experiments show that fine-tuned encoder-based models achieve strong performance, while classical ML ensembles handle label imbalance effectively topic classification task. These results establish evaluation metrics and reference points for structured municipal texts. To support responsible sharing and reproducible research, the dataset is openly released under FAIR principles and a Creative Commons Attribution-NonCommercial-NoDerivatives License, ensuring proper attribution, privacy protection, and copyright compliance. It is available through an open-access repository with persistent identifiers and rich metadata, enabling discovery and long-term availability. Future work will extend the dataset with deeper semantic layers (e.g., finer-grained personal information, subject summaries) and an English-translated version with aligned annotations to support cross-lingual exploration and comparative studies.

\section*{Limitations}\label{sec:limitations}

While CitiLink-Minutes provides a valuable resource of annotated municipal administrative documents, it is currently limited to six municipalities, which may restrict the applicability of pre-trained models to other municipalities. Another limitation concerns the recording of voting outcomes, which are sometimes available only at the party level, requiring external knowledge bases to map councilors. In addition, the current version of the voting layer in the annotation scheme captures only the main subject under discussion and does not yet support finer-grained annotation of sub-subjects. In cases where multiple votes are held on different aspects of the same subject, only the overarching subject is linked to the voting instance, while specific sub-subjects remain unrepresented. A further limitation concerns personal identifiers (PI), which are currently replaced with ***. Ideally, each PI would be annotated with its respective class, allowing for more advanced de-identification techniques; ongoing work is addressing this limitation. 

\section*{FAIR Principles}\label{sec:fair}
To support responsible sharing and reuse, we adhere to FAIR principles (Findable, Accessible, Interoperable, and Reusable)  \cite{wilkinson2016fair}. All data was collected in agreement with the participating municipalities, with strict attention to copyright and privacy. The dataset is published in an open-access repository with persistent identifiers \cite{citilink2025} and rich metadata, enabling discovery and long-term availability, under a Creative Commons Attribution-NonCommercial-NoDerivatives License, which ensures proper attribution and supports research use.


\subsubsection*{Preprint and Version of Record}
This preprint has not undergone peer review (when applicable) or any post-submission improvements or corrections. The Version of Record of this contribution is published in {Advances in Information Retrieval. ECIR 2026. Lecture Notes in Computer Science, vol 16486. Springer, Cham}, and is available online at https://doi.org/10.1007/978-3-032-21321-1\_56.

\subsubsection*{\ackname}
This work was funded within the scope of the project  CitiLink, with reference 2024.07509.IACDC, which is co-funded by Component 5 - Capitalization and Business Innovation, integrated in the Resilience Dimension of the Recovery and Resilience Plan within the scope of the Recovery and Resilience Mechanism (MRR) of the European Union (EU), framed in the Next Generation EU, for the period 2021 - 2026, measure RE-C05-i08.M04 - ``To support the launch of a programme of R\&D projects geared towards the development and implementation of advanced cybersecurity, artificial intelligence and data science systems in public administration, as well as a scientific training programme,'' as part of the funding contract signed between the Recovering Portugal Mission Structure (EMRP) and the FCT - Fundação para a Ciência e a Tecnologia, I.P. (Portuguese Foundation for Science and Technology), as intermediary beneficiary\footnote{\url{https://doi.org/10.54499/2024.07509.IACDC}}. We would also like to acknowledge Rodrigo Silva for managing the hosting of the dataset demo.

\bibliographystyle{splncs04}
\bibliography{mybibliography}

\end{document}